%% file: root.tex
\documentclass[letterpaper, 10 pt, conference]{ieeeconf}  %

\IEEEoverridecommandlockouts                              %

\usepackage{times}
\usepackage{epsfig}
\usepackage{graphicx}
\usepackage{svg}
\usepackage{amsmath}
\usepackage{amssymb}
\usepackage[noend]{algpseudocode}
\usepackage{balance}
\usepackage{url}
\usepackage{authblk}
\usepackage{multirow}
\usepackage{hyperref}
\hypersetup{
    colorlinks=true,
    linkcolor=blue,
    filecolor=magenta,      
    urlcolor=cyan,
    pdftitle={Overleaf Example},
    pdfpagemode=FullScreen,
    }

\newcommand{\eg}{\emph{e.g.},}
\newcommand{\ie}{\emph{i.e.},}

\def\secref#1{Section~\ref{#1}}
\def\figref#1{Figure~\ref{#1}}
\def\tabref#1{Table~\ref{#1}}
\def\eqref#1{Equation~(\ref{#1})}

\usepackage{cite}
\usepackage[font=small]{caption}
\usepackage{subcaption}
\usepackage{graphbox}
\usepackage{soul}

\usepackage{color, colortbl}
\usepackage{tabularx}

\usepackage[linesnumbered,ruled]{algorithm2e}
\usepackage[flushleft]{threeparttable}
\usepackage{lipsum}
\usepackage{amssymb}%
\usepackage{pifont}%
\newcommand{\cmark}{\ding{51}}%
\newcommand\numberthis[1][]{%
    \refstepcounter{equation}%
    \ifx#1\empty\else\label{eq:#1}\fi%
    \tag{\theequation}%
}

\makeatletter
\def\BState{\State\hskip-\ALG@thistlm}
\makeatother
\algdef{SE}[DOWHILE]{Do}{doWhile}{\algorithmicdo}[1]{\algorithmicwhile\ #1}

\graphicspath{ {figures/} }

\title{\LARGE \bf
{InCloud: Incremental Learning for Point Cloud Place Recognition}
}

\author{Joshua Knights$^{1,2}$, Peyman Moghadam$^{1,2}$, Milad Ramezani $^{1}$, Sridha Sridharan$^{2}$, Clinton Fookes$^{2}$   
\thanks{
$^1$ Joshua Knights, Peyman Moghadam and Milad Ramezani are with the Robotics and Autonomous Systems, DATA61, CSIRO, Brisbane, QLD 4069, Australia. 
E-mails: {\tt\footnotesize \emph{Joshua.Knights, Peyman.Moghadam, Milad Ramezani}@data61.csiro.au} }
\thanks{
$^{2}$ Joshua Knights, Peyman Moghadam, Sridha Sridharan, Clinton Fookes are with the research program in Signal Processing, Artificial Intelligence and Vision Technologies (SAIVT) at the Queensland University of Technology (QUT), Brisbane, Australia.
E-mails: {\tt\footnotesize \emph{joshua.knights, peyman.moghadam, s.sridharan, c.fookes}@qut.edu.au}}
} 

\usepackage{arydshln}
\newcolumntype{x}[1]{>{\centering\arraybackslash\hspace{0pt}}p{#1}}
\newcolumntype{M}[1]{>{\centering\arraybackslash}m{#1}}
\newcolumntype{L}[1]{>{\raggedright\arraybackslash} m{#1} }

\usepackage{titlesec}
\titlespacing*{\section}{0pt}{0mm}{0mm}

\begin{document}

\maketitle

\input{chapters/abstract}

\input{chapters/introduction-new}

\input{chapters/relatedworks}

\input{chapters/problemdef}

\input{chapters/methodology}

\input{chapters/experiments}

\input{chapters/results}

\input{chapters/conclusion}

{\small
        \bibliographystyle{IEEEtran}
        \bibliography{ref}
}

\end{document}

%% file: chapters/abstract.tex
\begin{abstract}

    Place recognition is a fundamental component of robotics, and has seen tremendous improvements through the use of deep learning models in recent years.  Networks can experience significant drops in performance when deployed in unseen or highly dynamic environments, and require additional training on the collected data.  However naively fine-tuning on new training distributions can cause severe degradation of performance on previously visited domains, a phenomenon known as catastrophic forgetting.  In this paper we address the problem of incremental learning for point cloud place recognition and introduce \textit{InCloud}, a structure-aware distillation-based approach which preserves the higher-order structure of the network's embedding space.  We introduce several challenging new benchmarks on four popular and large-scale LiDAR datasets (Oxford, MulRan, In-house and KITTI) showing broad improvements in point cloud place recognition performance over a variety of network architectures.  To the best of our knowledge, this work is the first to effectively apply incremental learning for point cloud place recognition.  Data pre-processing, training and evaluation code for this paper can be found at \href{https://github.com/csiro-robotics/InCloud}{https://github.com/csiro-robotics/InCloud}.

\end{abstract}

%% file: chapters/introduction-new.tex
\section{Introduction}
\label{sec:intro}

\begin{figure}[t]
    \centering
    \begin{subfigure}[b]{\linewidth}
        \centering 
        \includegraphics[width=0.45\textwidth]{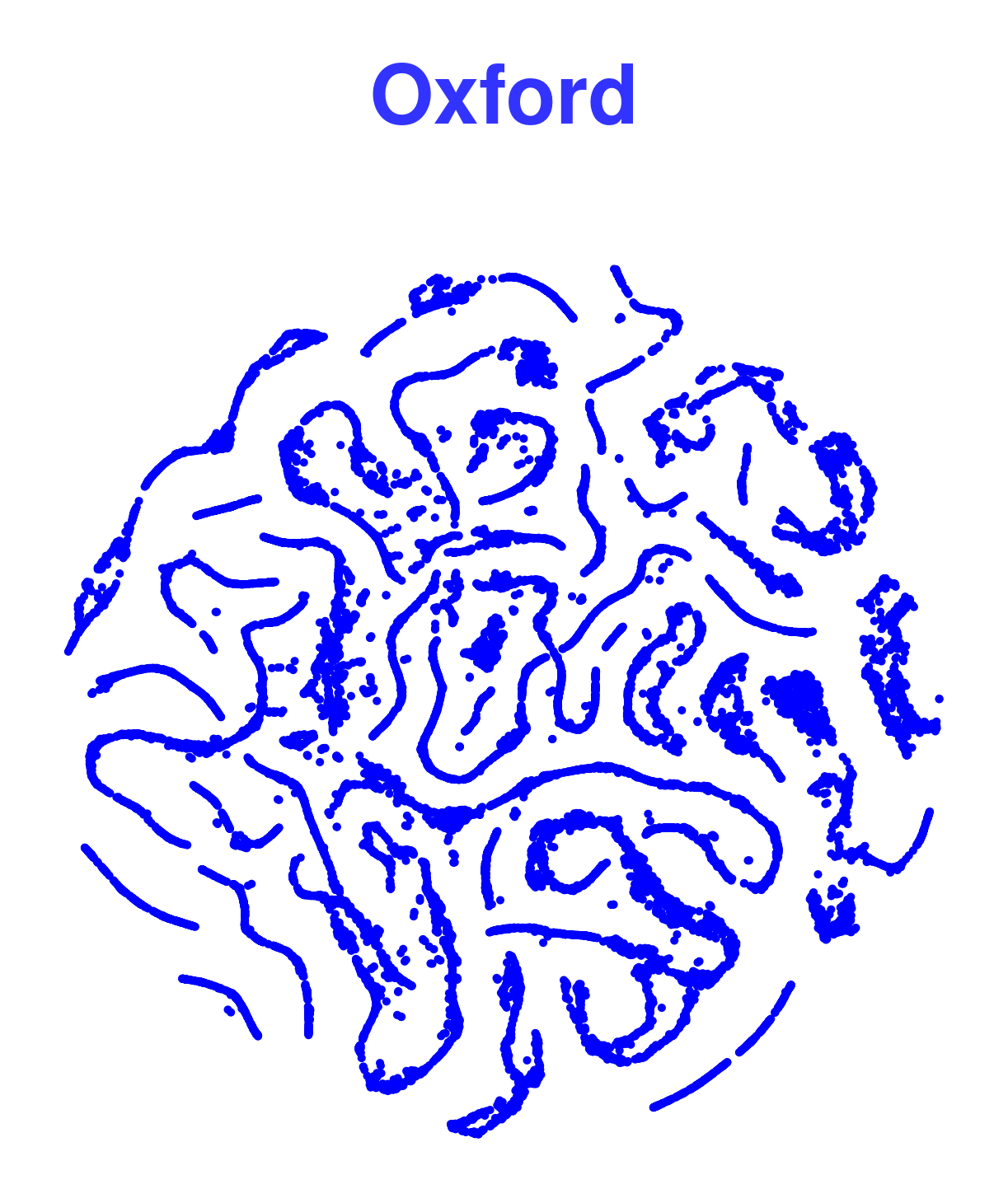}
        \includegraphics[width=0.45\textwidth]{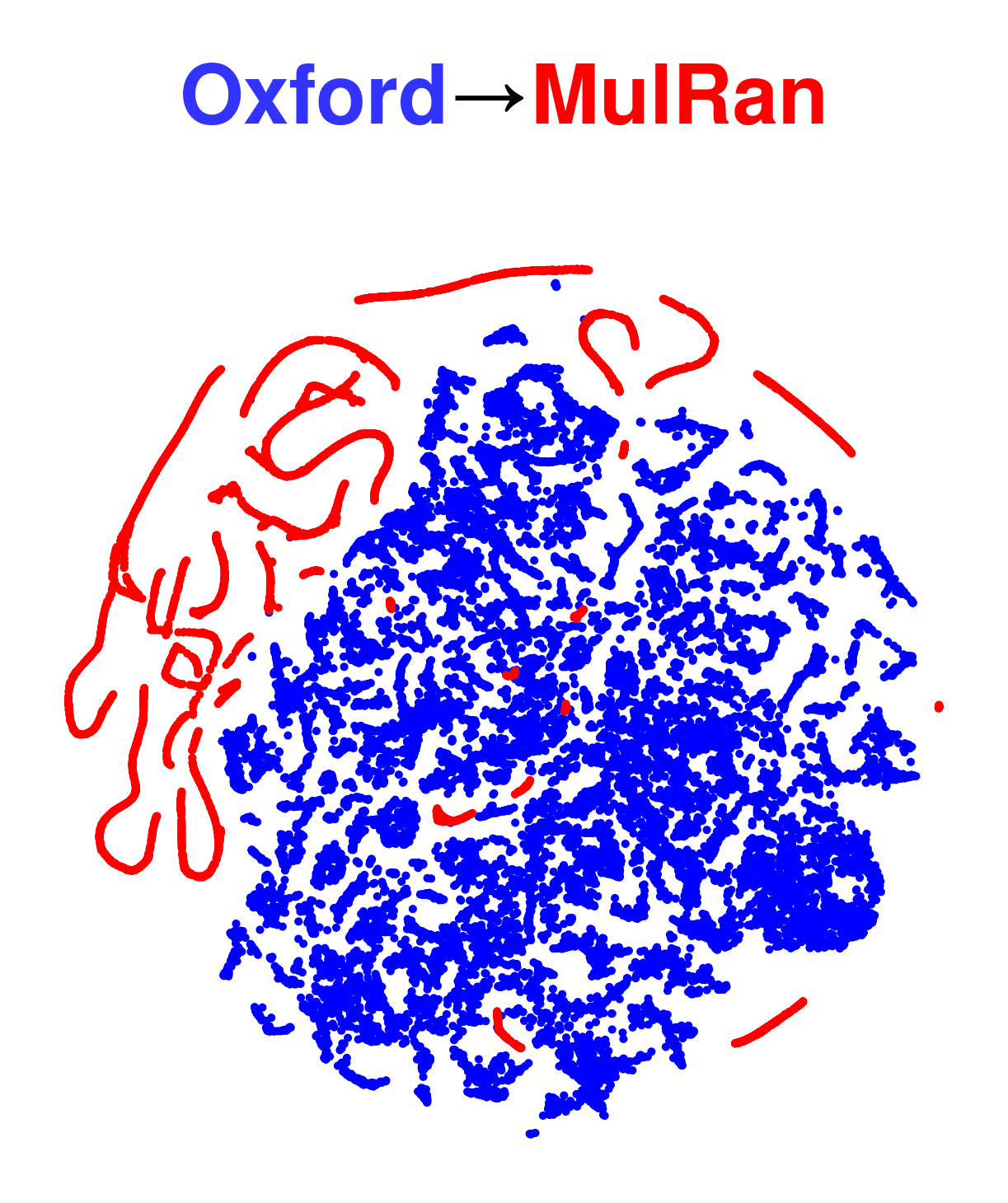}
        \caption{  Fine-tuning}
    \end{subfigure}
    
    \begin{subfigure}[b]{\linewidth}
        \centering 
        \includegraphics[width=0.45\textwidth]{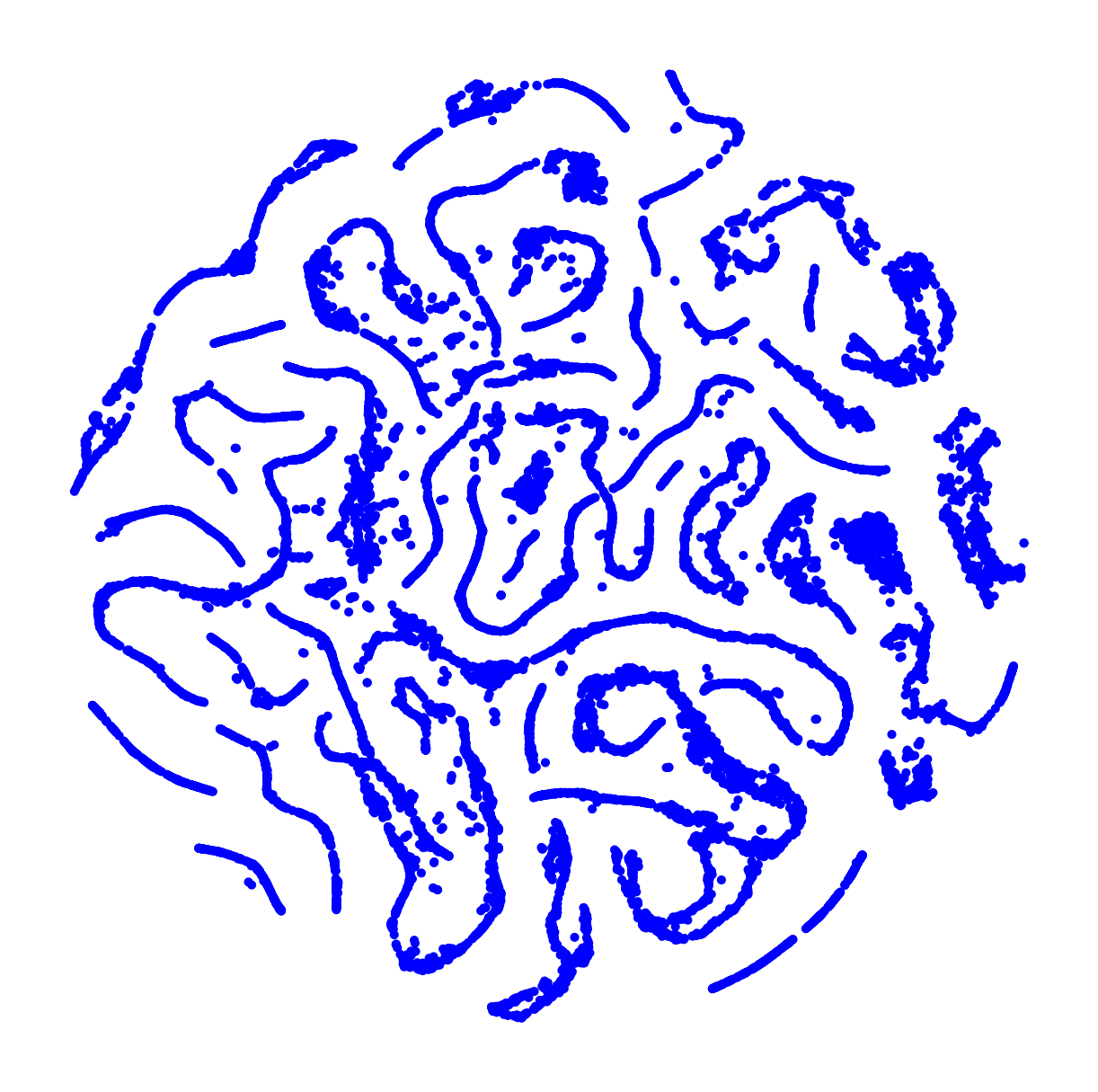}
        \includegraphics[width=0.45\textwidth]{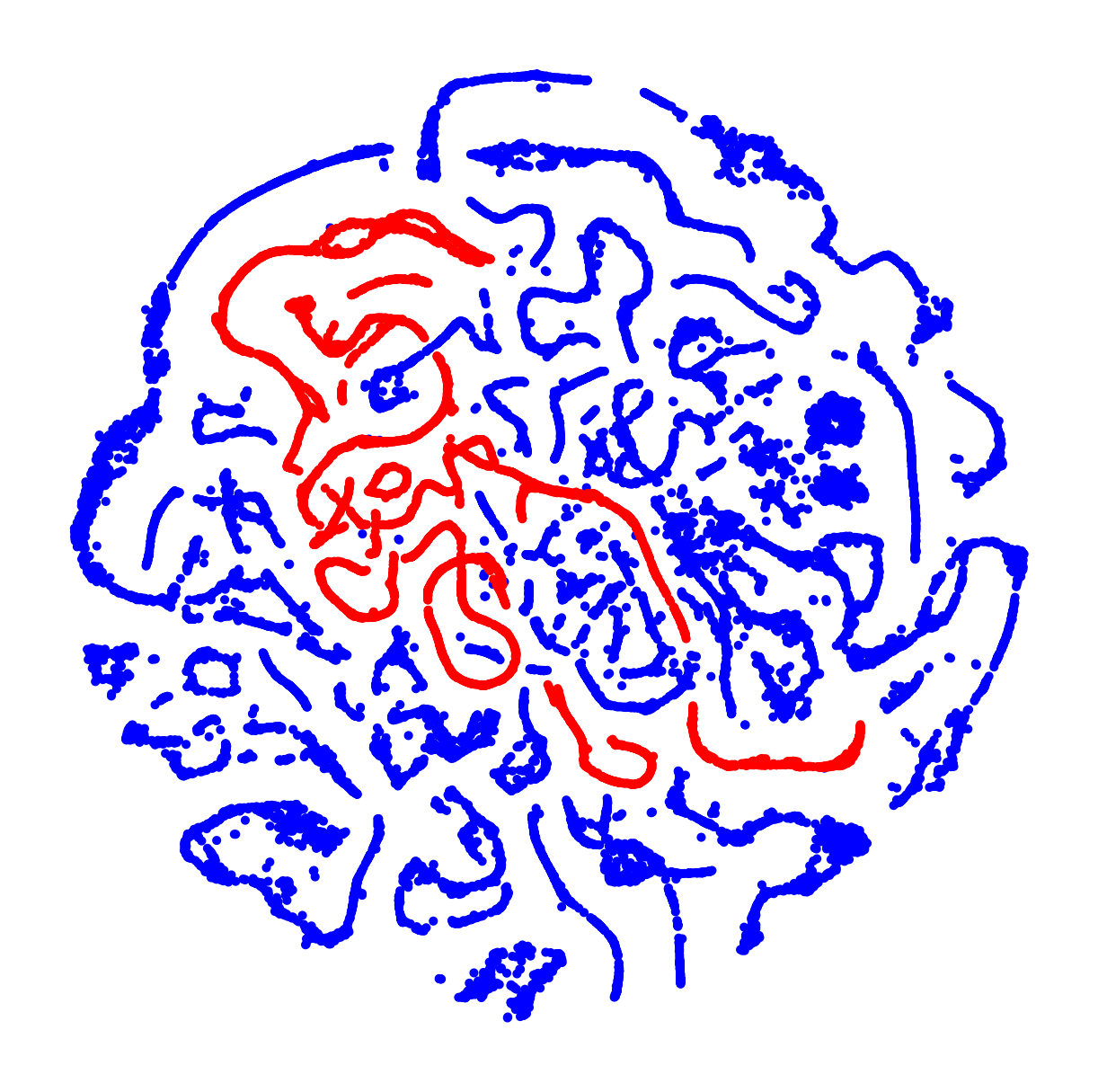}
        \caption{  \textit{InCloud} (ours)}
    \end{subfigure}

    \caption{  t-SNE of global descriptors of a point cloud place recognition network from Oxford (blue) to the DCC environment in MulRan (red). Fine-tuning (a) results in a collapse of the intricate structure of the embedding space required for place recognition on previously trained environments. Our approach \textit{InCloud} (b), preserves the structure of the embedding space between training environments.}
    \label{fig:hero}
    \vspace{-8mm}
\end{figure}
Place recognition is a fundamental component of robotics, and is essential for reliable loop closures in Simultaneous Localisation and Mapping (SLAM) or global relocalisation given a prior map in GPS-denied environments.

In recent years, there has been tremendous improvement in the accuracy of LiDAR place recognition methods using deep neural network models \cite{Uy2018PointNetVLADDP,vidanapathirana2021logg3d,komorowski2021minkloc3d}. 
The performance of these learning systems is often evaluated on large-scale LiDAR datasets based on the assumption that the training data is Identically and Independently Distributed (IID). Under this setting all the environments are available during training, and presumed to follow a stationary distribution. However, these assumptions are unrealistic for models deployed to environments which are highly dynamic or changing, in which case training on new data is required.  Legacy data which the model was previously trained on may be unrecorded, inaccessible, or infeasible to retain at scale on a platform due to storage constraints as the model learns over multiple environments.   %

In practical robotics applications, place recognition models should be able to continuously learn from new scenes without retraining from scratch while preserving knowledge about previously learned environments. This setup is referred to as Incremental Learning. Incremental Learning aims to dynamically update the model from sequentially arriving data streams while overcoming catastrophic forgetting, \ie{} degrading performance on past domains.  This is known as the neural stability-plasticity dilemma - maintaining strong performance on previous domains (neural \textit{stability}), while continuously adapting to the new domains (neural \textit{plasticity}) \cite{DeLange2021ACL}.

While incremental learning has recently been studied in computer vision for image classification \cite{Li2018LearningWF,Rebuffi2017iCaRLIC,Zhu2021PrototypeAA,Simon2021OnLT,Volpi2021ContinualAO}, object detection \cite{Perez-Rua_2020_CVPR}  and semantic segmentation \cite{Cermelli2020ModelingTB,Maracani2021RECALLRC,Douillard2021PLOPLW}, no attention has been devoted to incremental learning for point cloud place recognition in the robotics community.  Hence, in this paper we are the first to introduce incremental learning for LiDAR-based point cloud  place recognition, and effectively address (as shown in \figref{fig:hero}) the problem of catastrophic forgetting caused by learning in new domains which negatively impacts the embedding spaces learned from past training. 

To alleviate forgetting we propose \textit{InCloud}, which is the first approach to use structure aware knowledge distillation for point cloud place recognition. In particular, \textit{InCloud} distills the angular relationship between embedded triplets rather than directly distilling the output logit used in classical distillation-based frameworks. This encourages the network to preserve the high-level structure of the embedding space while retaining the necessary flexibility to learn from training data in new domains. 
We also propose a relaxation of the distillation constraint to encourage a gradual transition of the network parameters between training steps to achieve a stronger solution over multiple training domains.

To summarise, the main contributions of this paper are as follows: 

\begin{itemize}
\item We are the first to introduce the task of incremental learning for point cloud place recognition, in particular effectively addressing the problem of collapse of the intricate structure of the old embedding space as the network adjusts to the new environment.   
\item We propose \textit{InCloud}, which encourages the network to preserve the structure of its embedding space using an angular distillation loss and employs a gradual relaxation of the distillation constraint to improve the stability-plasticity trade-off during training.  
\item We establish several new  challenging benchmarks on four popular large-scale, outdoor LiDAR datasets to analyse the performance of incremental learning methods for point cloud place recognition. 
\end{itemize}

%% file: chapters/relatedworks.tex
\section{Related Work}
\label{sec:relwork}
\subsection{Point Cloud Place Recognition}
\label{sec:relworkpr}

{There is a large body of work related to LiDAR-based place recognition in robotics~\cite{himstedt2014large,he2016m2dp}. Handcrafted approaches such as SHOT~\cite{salti2014shot}, ScanContext~\cite{kim2018scan} and ScanContext++~\cite{gskim-2021-tro} have been designed to either encode the entire point cloud into a global descriptor in an end-to-end fashion or extract local features. Despite the fact these approaches have shown rotational invariance regardless of viewpoint changes, their discriminative power remains limited in multi-session scenarios or cluttered environments with dynamic entities.}

PointNetVLAD~\cite{Uy2018PointNetVLADDP} is a seminal work in deep learning based large-scale point cloud place recognition.  It employs PointNet \cite{qi2017pointnet}, a rudimentary point cloud feature extractor which leverages symmetric max pooling to obtain latent features invariant to permutation of the input cloud, in conjunction with NetVLAD \cite{Arandjelovi2018NetVLADCA}, a deep-learning based aggregator which converts the coarse local features into a global descriptor.  %

MinkLoc3D~\cite{komorowski2021minkloc3d} uses a sparse-voxel convolutional backbone for its feature extractor, which voxelises the input point cloud and performs 3D convolution over only the activated regions of the voxel tensor to reduce training time and memory consumption.  GeM \cite{GeM2017} pooling is then used to convert coarse local features into a robust global descriptor.  

Several other recent approaches have explored using higher-order pooling to produce the network's global descriptor.  Locus~ \cite{vidanapathirana2021locusLP} uses a pre-trained network to extract segment features from sequential point clouds, and uses second-order pooling to create a temporally aware global descriptor.  More recently, LoGG3D-Net \cite{vidanapathirana2021logg3d} uses a sparse-voxel network to extract local features for each input point and then uses second-order pooling with differentiable Eigen-value power normalisation to obtain a global descriptor that is invariant to the permutation of the local features, as well as enforcing consistency between registered local features across adjacent point clouds.  {Finally, several methods have recently been proposed which incorporate self-attention into their network architectures \cite{Hui_2021_ICCV,Zhou2021NDTTransformerL3} for feature extraction, or perform place recognition in tandem to predicting the orientation of the platform \cite{Chen2022OverlapNetAS,Xu2021DiSCODS}.}

While all these methods focus on producing a robust, viewpoint-invariant global descriptor for point cloud place recognition, when learning from a new domain they must either train in conjunction with all previously accessed training data or fine-tune solely on the newly presented data.  The former scenario is often unrealistic in a robotics context due to legacy data being unrecorded, inaccessible for ethical or proprietary reasons, or being too cumbersome to retain on disk.  Conversely, fine-tuning causes the network to forget key knowledge in previously visited domains, with a corresponding drop in performance.

\subsection{Incremental Learning}
Incremental Learning (also known as lifelong \cite{gao2021airloop}, continual \cite{Michieli2021ContinualSS} or sequential \cite{DeLange2021ACL} learning) aims to dynamically update the model from sequentially arriving data streams while overcoming catastrophic forgetting, \ie{} degrading performance on the old domains.
A key challenge in incremental learning is to minimise catastrophic forgetting, leveraging new information to update the model while preserving what has been learned in the past.  Incremental learning approaches can broadly be categorised into one of two groups: rehearsal \cite{Rebuffi2017iCaRLIC,Chaudhry2018RiemannianWF} or pseudo-rehearsal \cite{Odena2017ConditionalIS,Zhu2021PrototypeAA} based methods, which use either stored exemplars or generative models to replay data from previous tasks, or regularisation of the weights \cite{Kirkpatrick2017OvercomingCF,Chaudhry2018RiemannianWF}, features \cite{douillard2020podnet,Douillard2021PLOPLW} and outputs \cite{Li2018LearningWF,Simon2021OnLT,Cermelli2020ModelingTB} to preserve knowledge between training tasks.  

 {However, many of the existing incremental learning approaches for classification tasks are either directly inapplicable due to a reliance on priors such as semantic class clusters} \cite{Rebuffi2017iCaRLIC,Douillard2021PLOPLW,Maracani2021RECALLRC} {or recent-class bias} \cite{Hou_2019_CVPR,dong2017class}, {or poorly suited due to distilling individual input embeddings rather than the overall embedding structure.}  Concurrent to this submission, Airloop \cite{gao2021airloop} proposed a regularisation-based lifelong learning method for visual loop closure detection using a  euclidean-distance knowledge distillation loss on images. In contrast to this work, we use a higher-order angular distillation loss to preserve the structure of the embedding space between training steps while retaining flexibility to learn from new training data by introducing distillation relaxation.

%% file: chapters/problemdef.tex
\section{Problem Setup}
\label{sec:problem_setup}
In this section we formalise the problem of incremental learning for LiDAR point cloud place recognition, and briefly discuss the major obstacles to be addressed in extending current approaches to an incremental setting.
\subsubsection{Point Cloud Place Recognition}
For completeness, we start by defining conventional point cloud place recognition.  Given an input point cloud $\mathbf{P} = \left\{p_i \in \mathbb{R}^3 | i = 1..N\right\}$, our network with parameters $\theta$ encodes the point cloud into a global descriptor $v \in \mathbb{R}^d$ with $d$ feature channels.  Consequently, given a training set of $M$ point clouds $\mathbf{S} = \left\{\left(\mathbf{P}_j, l_j\right) | j = 1..M\right\}$ consisting of point clouds and their corresponding global locations $l = \left(x,y\right)$, we try to learn the network parameters $\theta$ such that for any subset of samples $\left\{\left(\mathbf{P}_a, l_a\right),\left(\mathbf{P}_b, l_b\right),\left(\mathbf{P}_c, l_c\right) \right\}$:
\begin{equation}
    \label{eq:pr}
    \mathcal{D}\left(l_a,l_b\right) \leq \mathcal{D}\left(l_a,l_c\right) \longrightarrow \left\|v_a-v_b\right\|_2 \leq \left\|v_a - v_c\right\|_2 , 
\end{equation}
where, $\mathcal{D}$ denotes geometric distance in the real world and $\left\|\cdot\right\|_2$ denotes the $L_2$ distance in the embedding space between global descriptors.  Learning the parameters $\theta$ that addresses the above problem is generally performed using a metric learning solution by applying a loss function $\mathcal{L}$ which acts on the global descriptors extracted from a tuple of training samples.

\subsubsection{Incremental Point Cloud Place Recognition}
In the incremental setting we consider a training dataset $\mathbf{S} = \left\{\mathbf{S}^{t}\right\}^T_{t=1}$ which consists of $T$ disjoint training domains, where every domain represents a distinct and disjoint training environment.  We also assume that we have access to a small memory set $\mathbf{E}^{t}$ such that it is a subset of $\mathbf{S}^{1:t}$, where $\mathbf{S}^{1:t} = \bigcup_{i=1}^t \mathbf{S}^{i}$.  At each incremental training step $t$ the network is initialised with the trained parameters from the previous step, $\theta^{t-1}$, and is trained on $\hat{\mathbf{S}}^{t} = \mathbf{S}^{t} \cup \mathbf{E}^{t}$, with the goal of satisfying \eqref{eq:pr} on the domain of $\mathbf{S}^{t}$. 

However, inter-domain performance for networks is often limited by the network overfitting to the currently available training data.  Existing point cloud place recognition approaches have no mechanism to retain salient information from previous training steps, which results in degradation of performance between domains as the network repeatedly overfits to the new training set.  Conversely, unduly strict constraints to prevent forgetting can severely impact the network's plasticity, causing intransigence: the inhibition of the network's ability to learn on new domains.  Our approach uses a structure-aware knowledge distillation loss to regularise the network output, and a novel distillation relaxation method to gradually reduce the regularisation constraint when training on new domains, allowing for a smoother transition of the network parameters between training steps.%

%% file: chapters/methodology.tex
\section{InCloud: \underline{In}cremental  Point \underline{Cloud} Place Recognition}
\label{sec:incloud}

\begin{figure}[t]
    \centering
    \includegraphics[width=0.8\linewidth]{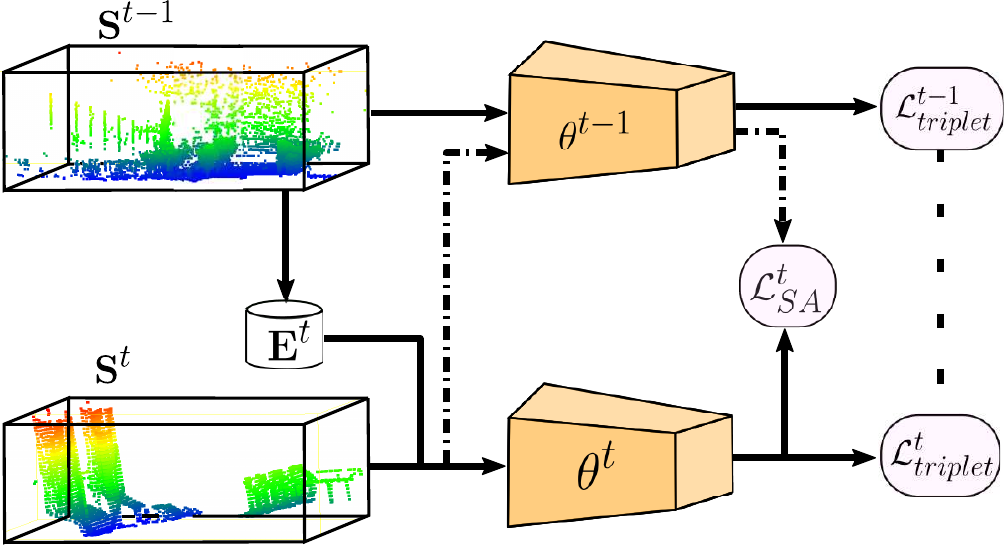}
    \caption{  Overview of \textit{InCloud}.  At step $t$, the model is trained with a combination of data from the current scene $\mathbf{S}^t$ and a memory set containing exemplars from previous scenes $\mathbf{E}^t$.  Training data is fed to both the currently training model {(solid arrows)} and a frozen model {(dashed arrows)} from the last incremental training step to produce a metric loss $\mathcal{L}^t_{triplet}$, which learns from the current data distribution, and $\mathcal{L}^t_{SA}$ which preserves the structure of the embedding space between steps.}
    \label{fig:diagram}
    \vspace{-5mm}
\end{figure}

In this section we propose an incremental learning approach for point cloud place recognition called \textit{InCloud}, shown in \figref{fig:diagram}.

\subsection{Metric Learning for Place Recognition}
\label{sec:triplet}
To train our network on the place recognition task, we use a triplet margin loss:
\begin{equation}
    \mathcal{L}^t_{triplet} = \max\left\{d\left(v^t_a, v^t_p\right) - d\left(v^t_a,v^t_n\right) + \delta, 0\right\},
    \label{eq:triplet}
\end{equation}
where, $d\left(x,y\right) = \left\|x - y\right\|_2$ is the $L_2$ distance between two embeddings $x$ and $y$, $v^t_a,v^t_p,v^t_n$ are the embeddings at training step $t$ of a training triplet formed from an anchor point cloud $\mathbf{P}_a$ and a corresponding positive and negative point cloud $\mathbf{P}_p$ \& $\mathbf{P}_n$, and $\delta$ is a margin hyper-parameter.
\subsection{Structure-Aware Knowledge Distillation}
\label{sec:lsa}

\eqref{eq:pr} tells us that place recognition performance is largely invariant to the absolute position of global representations in the embedding space, with the \textit{relationship} between representations being the key to strong performance.  Consequently distilling the absolute position of a given point cloud's embedding between incremental training steps presents a major obstacle to the network's plasticity, as it imposes several constraints which limit the available space for the network to learn from new data while contributing very little to preserving the embedding structure. 

\begin{figure}[t]
    \centering
    \includegraphics[width=1.0\linewidth]{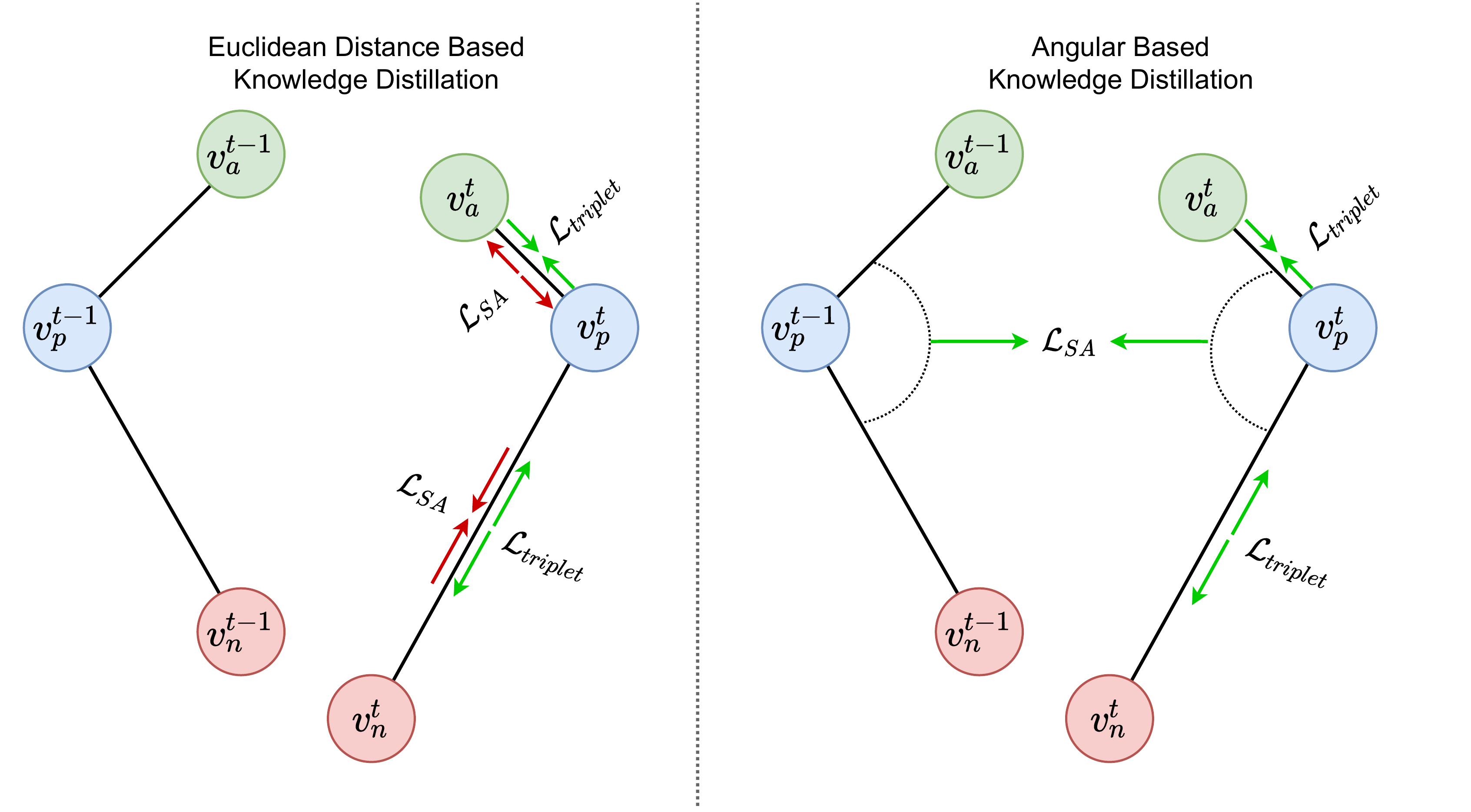}
    \caption{  {Euclidean distance vs. angular-based knowledge distillation.  The learning signal from distilling euclidean distance can conflict with the learning signal from the triplet loss when positive or negative pairs are closer or further apart respectively than in the frozen model from a previously visited environment, inhibiting training.  The angular-based distillation is invariant to euclidean distance, which allows the triplet loss to scale distance between features appropriately while preserving the higher-order structure of the embedding space.  Solid and dashed lines represent the structure of the frozen model from a previous environment and the currently training model, respectively.}}
    \label{fig:higher-order}
    \vspace{-5mm}
\end{figure}

{Previous works} \cite{gao2021airloop,ChenFeat2021} {have distilled the euclidean distance between embeddings to address this issue; however, as shown in \figref{fig:higher-order} euclidean distance based distillation can conflict with the training signal from the triplet loss when positive and negative pairs of embeddings are closer to or further apart from each other than in the frozen teacher model.  On these grounds we introduce a structure-aware knowledge distillation loss which distills the angular relationship between tuples of embeddings.  As a higher-order structural property the angular distillation is invariant to rotation, translation and scaling of the tuple in the embedding space, preserving the relative structure of the embedding while reducing the potential conflict between the two learning signals.}

The angle formed by a triplet of embeddings in the feature space can be measured as:
\begin{equation}
        \phi^t_{i,j,k} = \left<\frac{v^t_i-v^t_j}{\left\|v^t_i-v^t_j\right\|_2},\frac{v^t_k-v^t_j}{\left\|v^t_k-v^t_j\right\|_2}\right>,
        \label{eq:angle}
\end{equation}
where $\left<\cdot,\cdot\right>$ is the inner product and $\phi^{t}_{i,j,k}$ is the angle formed by a tuple of embeddings $v_i^t,v_j^t,v_k^t$ in a training batch $B$.  From this we now define our structure-aware loss ($\mathcal{L}_{SA}$) as:
\begin{equation}
    \mathcal{L}^t_{SA} = \sum_{\left(i,j,k\right) \in B}\max\left\{h\left(\phi^{t-1}_{i,j,k} - \phi^{t}_{i,j,k}\right) - \kappa, 0\right\},
\end{equation}
where $h$ is the Huber loss which is defined as:
\begin{equation}
    h\left(x,y\right) = 
        \begin{cases}
            \frac{1}{2}\left(x-y\right)^2 & \mathrm{for} \left|x-y\right| \leq 1 \\
            \left(\left|x-y\right| - \frac{1}{2}\right), & \mathrm{otherwise}
        \end{cases}
\end{equation}
and $\kappa$ is a margin hyper-parameter.  The margin term encourages flexibility during training by preventing penalisation of the network output once the difference in angular relationship between old and new embeddings is constrained under a set threshold, leaving room for the model to adapt to new training data.

\subsection{Memory Construction}
The incremental memory $\mathbf{E}^{t}$ is a high-level abstraction to human memory in the vision perceptual process, storing a small number of salient examples to improve recall.  For categorisation-based tasks such as classification or segmentation, the standard approach to constructing the memory is to store a small set of selected training exemplars alongside their ground truth.  This approach is not possible for large-scale place recognition, as inputs do not have classes or other independent ground truths; success is defined by similarity and dissimilarity to other point clouds in the embedding space.

Therefore, we construct our incremental memory out of $R$ pairs of corresponding point clouds such that $\mathbf{E}^{t} = \left\{\left(\mathbf{P}_{r}, \mathbf{P}_{r_p}\right)\right\}_{r=1}^R$, where $\mathbf{P}_r$ is a point cloud from a previous training domain and $\mathbf{P}_{r_p}$ is a randomly selected positive pair of $\mathbf{P}_r$.  The constructed memory bank consequently contains $2R$ unique point clouds and ensures that the network is always able to form positive pairs when constructing triplets during training.

Point clouds in the memory are selected through random sampling  {after each training environment}, and when updating the memory bank we replace stored exemplars uniformly to ensure a balance between each previously visited domain.

\subsection{Distillation Relaxation}
\label{sec:relax}
The hyper-parameter $\lambda_{SA}$ which is used to weight the loss term $\mathcal{L}^t_{SA}$ is a major contributor to the stability-plasticity trade-off during the network's training, with higher magnitudes prioritising stability over plasticity and the inverse for lower magnitudes.  Instead of holding the value of $\lambda_{SA}$ constant we propose a distillation relaxation approach to gradually lower the magnitude of $\lambda_{SA}$ over the course of the network's training.  By doing so we reduce the disruption the network experiences when exposed to a new domain, allowing for a smoother transition of the network parameters between training steps and producing a stronger representation over all seen domains.  We use a sigmoid function to perform this relaxation as:
\begin{equation}
    \label{eq:schedule}
    \omega\left(\gamma\right) = \frac{1}{1 + e^{10\left({\gamma}/{\tau} - 0.5\right)}},
\end{equation}
where $\gamma$ is the current training epoch, $\tau$ is the total number of training epochs and $\omega$ represents the loss relaxation at any given training epoch.  Our adjusted weight $\lambda^\gamma_{SA}$ can then be expressed as:
\begin{equation}
    \mathcal{\lambda}_{SA}^{\gamma} = \omega\left(\gamma\right)\cdot \lambda_{init},
\end{equation}
where $\lambda_{init}$ is the initial value of the knowledge distillation weight at the start of training, and is a tunable hyper-parameter.

\subsection{Final loss}

Our overall objective now has the form of:
\begin{equation}
    \mathcal{L}^t = \mathcal{L}^t_{triplet} + \lambda_{SA}^{\gamma}\mathcal{L}^t_{SA}, 
\end{equation}
where $\mathcal{L}^t_{triplet}$ is the triplet loss, $\mathcal{L}^t_{SA}$ is the structure-aware loss, and $\lambda^{\gamma}_{SA}$ is the distillation relaxation.

%% file: chapters/experiments.tex
\section{Experimental Setup}

In this section we detail our experimental setup for the task of incremental learning on point cloud place recognition.  %
\subsection{Datasets}
\label{sec:datasets}

Models experience catastrophic forgetting when trained sequentially on datasets with dissimilar data distributions, which in our case can result from environmental changes (\eg{} different cities, regions) or changes in how the data was collected (\eg{} different LiDAR sensors).  To study the impact of catastrophic forgetting in point cloud place recognition and evaluate the performance of \textit{InCloud} we use four large-scale point cloud datasets: Oxford RobotCar~\cite{RobotCarDatasetIJRR}, In-house~\cite{Uy2018PointNetVLADDP}, MulRan\cite{Kim2020MulRanMR} and KITTI \cite{Geiger2013VisionMR}.
Table~\ref{tab:data} outlines the sensor types, train and test set sizes of the various environments used in our incremental learning setup.  To maintain input consistency, we pre-process all training and evaluation datasets as described for the Oxford dataset.

\textbf{Oxford: }Oxford RobotCar dataset~\cite{RobotCarDatasetIJRR} was collected using a SICK LMS-151 2D LiDAR scanner around the city of Oxford.  Based on the LiDAR scans, a database of submaps is built and each submap has the non-informative ground plane removed, and is downsampled to 4096 points with coordinates normalised between [-1,1].  During training, point clouds are considered positive matches if they are within 10m of each other and negative matches if more than 50m apart.  During testing, we regard a retrieved point cloud as a positive match if it is within 25m of the query scan.  We follow the standard convention used by \cite{Uy2018PointNetVLADDP} when creating our training and testing splits.

\textbf{In-house: }In-house dataset \cite{Uy2018PointNetVLADDP} is composed of five traversals each of three routes collected around the city of Singapore using a Velodyne HDL-64E LiDAR scanner.  We follow the standard convention for training and testing splits, employing the same positive and negative thresholds as the Oxford dataset.  %

\begin{table}[t]
    \centering
    \begin{tabularx}{\linewidth}{lXcc}
         \hline 
        Dataset & LiDAR Sensor & Train  & Test  \\%& Positive Radius & Negative Radius \\
         \hline 
         Oxford \cite{RobotCarDatasetIJRR} & SICK LMS-151  & 21711 & 3030 \\%& $<10m$ & $>25m$\\
         In-house \cite{Uy2018PointNetVLADDP} & Velodyne HDL-64E  & 6671 & 1766 \\%&$<10m$ & $>25m$\\
         DCC \cite{Kim2020MulRanMR} & Ouster OS1-64 & 5542 & 15039 \\%&$<10m$ & $>20m$\\
         Riverside \cite{Kim2020MulRanMR} & Ouster OS1-64 & 5537 & 18633 \\%&$<10m$ & $>20m$\\
         KITTI \cite{Geiger2013VisionMR} & Velodyne HDL-64E & - & 4542 \\%&$<3m$ & $-$\\
         \hline 
    \end{tabularx}
    \caption{  Train-Test splits in our incremental learning setup.}
    \label{tab:data}
    \vspace{-5mm}
\end{table}

\textbf{MulRan: }MulRan dataset~\cite{Kim2020MulRanMR}  contains scans collected from an Ouster OS1-64 sensor from multiple environments in South Korea. The dataset contains 12 sequences on four environments, of which we use the six sequences from the DCC and Riverside environments for training and testing.
During training, point clouds are considered positive matches if they are within 10m of each other and negative matches if more than 20m apart.  During testing, we regard a retrieved point cloud as a positive match if it is within 10m of the query.  
{For both environments we train on sequence $01$ and evaluate on sequences $02$ and $03$.}

{\textbf{KITTI: }KITTI odometry dataset }\cite{Geiger2013VisionMR} {contains 11 sequences of Velodyne HDL-64E LiDAR scans collected in Karlsruhe, Germany.  Following} \cite{Chen2022OverlapNetAS,Schaupp2019OREOSOR,gskim-2021-tro} {we use sequence 00 for evaluation in the \textit{Zero-Shot} protocol outlined below.} 
During testing, we regard a retrieved point cloud as a positive match if it is within 3m of the query.  %

We establish three experimental protocols to analyse the performance of incremental learning methods for point cloud place recognition. In the first protocol we perform two learning steps (\textit{2-Step}), after the first training set (Oxford), we add the following 3 sets (DCC, Riverside and In-house) all at once in the second incremental step. 
In the second protocol (\textit{4-Step}), we add 4 training sets sequentially, one by one in the following incremental order \textit{Oxford} $\rightarrow$ \textit{DCC} $\rightarrow$ \textit{Riverside} $\rightarrow$ \textit{In-house}. The \textit{4-Step} protocol is much more challenging due to its higher number of incremental steps.  {Finally, we hypothesise that a successful incremental learning method should also result in a most robust global embedding which generalises well to domains that have not been previously seen by the network during training.  Therefore we introduce a \textit{Zero-Shot} evaluation protocol which reports performance on KITTI sequence 00 after training on the \textit{4-Step} training protocol, which notably does not contain any environments from the KITTI dataset during training.}

\begin{figure*}[t]
     \centering
     \begin{subfigure}[b]{0.2\textwidth}
         \centering
         \includegraphics[width=\textwidth]{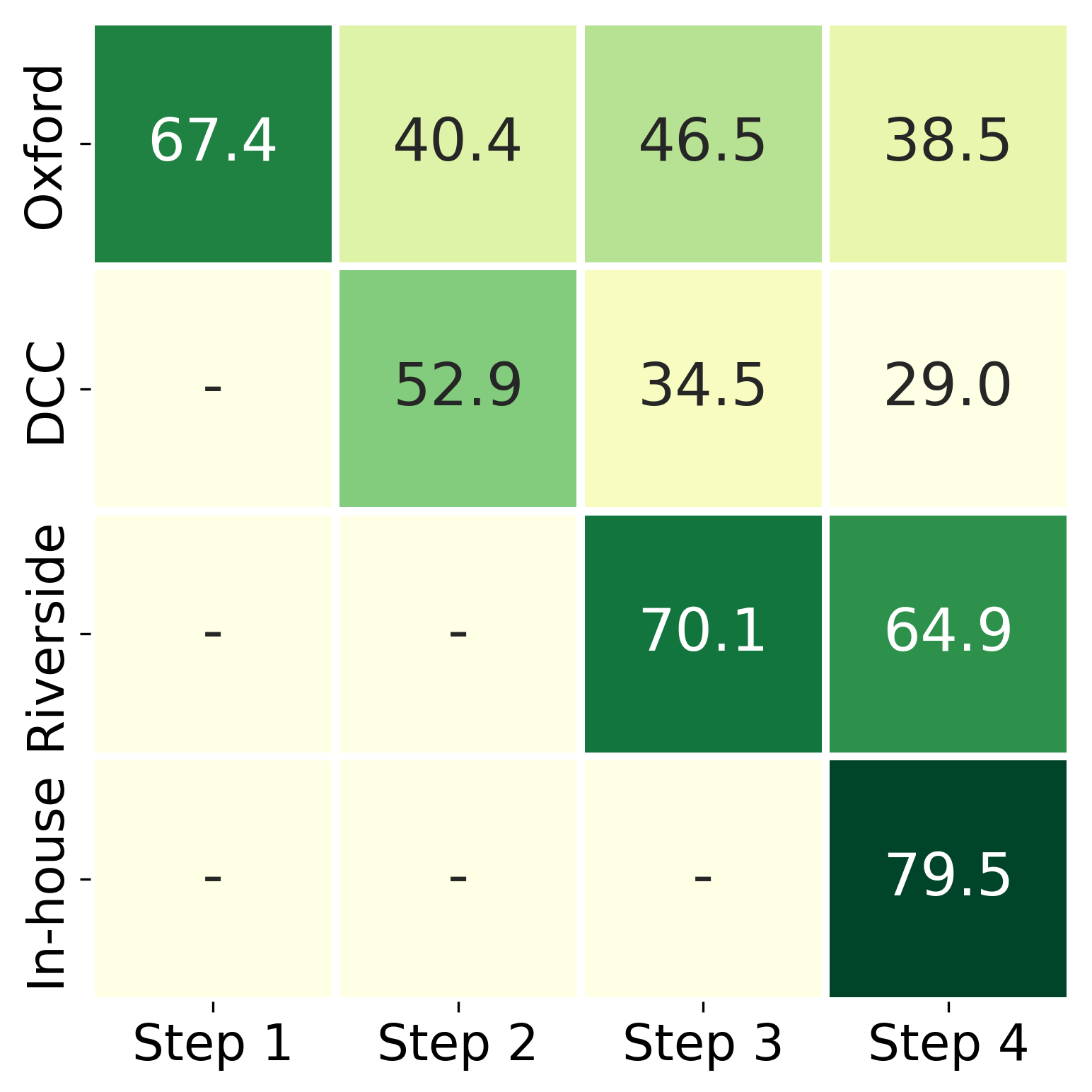}
         \caption{  PointNetVLAD~\cite{Uy2018PointNetVLADDP}}
     \end{subfigure}
     \hfill
     \begin{subfigure}[b]{0.2\textwidth}
         \centering
         \includegraphics[width=\textwidth]{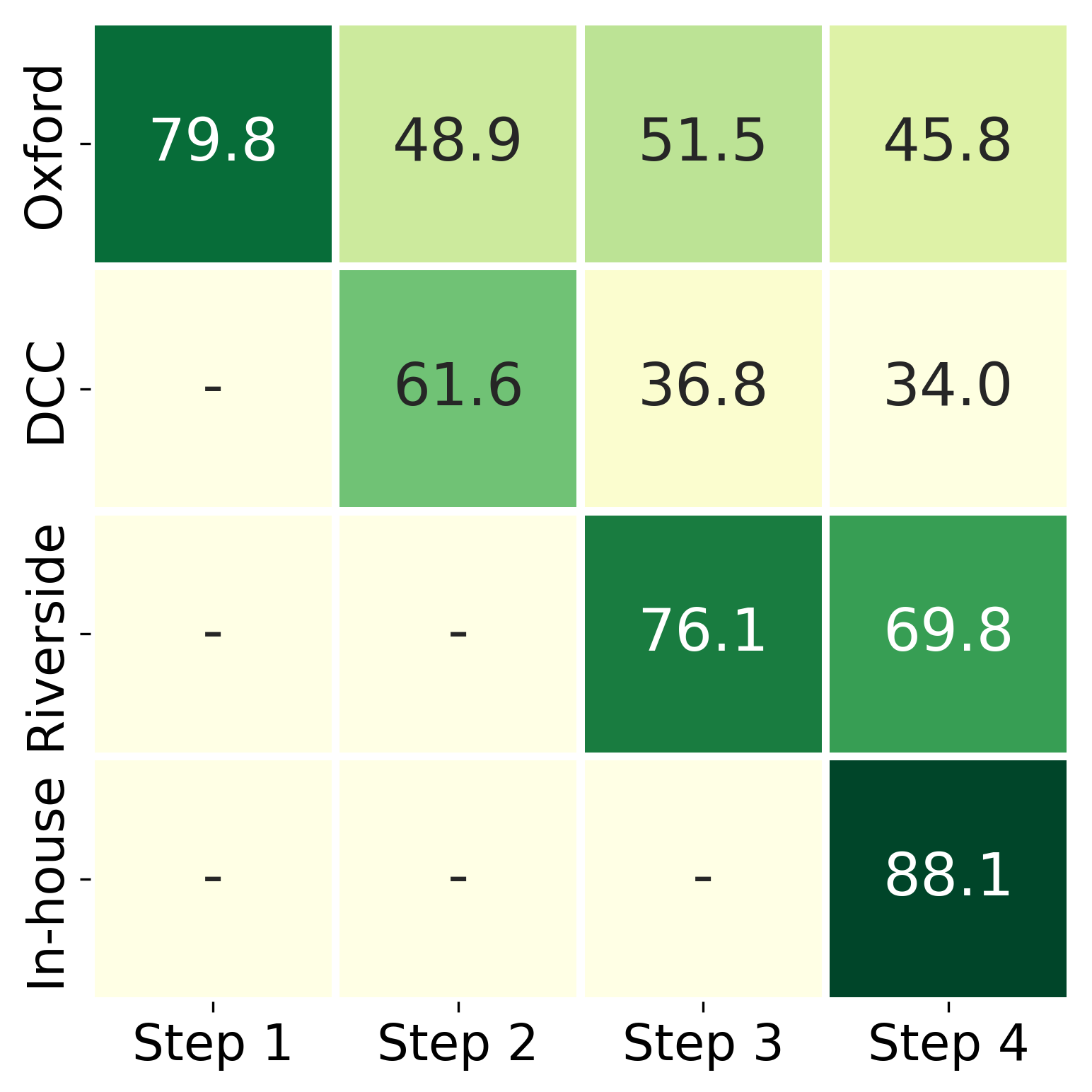}
         \caption{  LoGG3D-Net$\dagger{}$~\cite{vidanapathirana2021logg3d}}
     \end{subfigure}
     \hfill
     \begin{subfigure}[b]{0.2\textwidth}
         \centering
         \includegraphics[width=\textwidth]{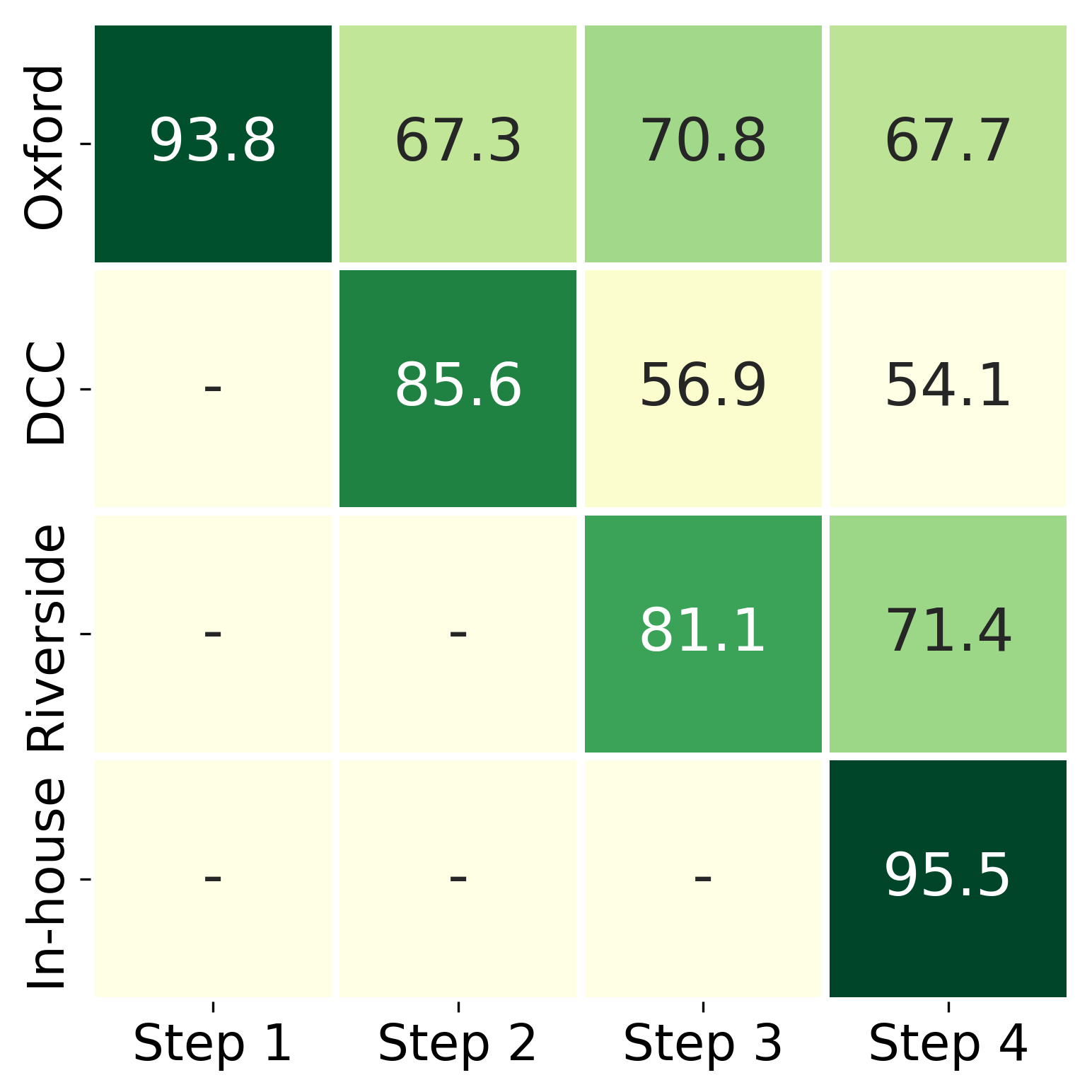}
         \caption{  MinkLoc3D~\cite{komorowski2021minkloc3d}}
     \end{subfigure}
        \caption{  The impact of catastrophic forgetting in the  \textit{4-Step} incremental protocol, across (a) PointNetVLAD \cite{Uy2018PointNetVLADDP} (b) LoGG3D-Net$\dagger{}$\cite{vidanapathirana2021logg3d} and (c) MinkLoc3D \cite{komorowski2021minkloc3d}.  Results in each cell represent Recall@1 on an environment at each training step.  The network achieves highest performance on any given environment in the training step it is introduced, after which catastrophic forgetting causes the performance to degrade.}
        \label{fig:forgetting}
        \vspace{-5mm}
\end{figure*}

\subsection{Network Architectures}
\label{sec:architectures}

To demonstrate that \textit{InCloud} is agnostic to network architecture, we use three state-of-the-art network architectures: PointNetVLAD~\cite{Uy2018PointNetVLADDP}, LoGG3D-Net~\cite{vidanapathirana2021logg3d} and MinkLoc3D~\cite{komorowski2021minkloc3d}. {These three networks are chosen as their architectures represent a significant cross-section of the different styles of feature extractors and feature aggregators used in the literature (see \secref{sec:relworkpr} for more details).} We found registering the In-house dataset was not feasible due to insufficient overlap between adjacent point clouds. Therefore, we use the abbreviation LoGG3D-Net$\dagger{}$ to indicate the point-based local consistency loss in LoGG3D-Net was not used during training.

For the first incremental step, we train the network from randomly initialised parameters.  In each following incremental step, we train the model for 60 epochs with an initial learning rate of $1e^{-3}$ reduced to $1e^{-4}$ at 30 epochs.  To increase the speed of training, we perform in-batch hard negative mining as proposed in \cite{komorowski2021minkloc3d}.  We used ADAM as the optimiser with a weight decay of $1e^{-3}$, and set our incremental memory size to $K = 256$ during training.  Random rotation and flipping of input 3D point clouds are used as data augmentation methods during training.  %

\subsection{Evaluation Metrics}
\label{sec:metrics}
To evaluate the performance of our incremental training setting, we report the mean Recall@1 (mR@1) across all training environments.  We also adopt an additional metric introduced by \cite{Chaudhry2018RiemannianWF}: Forgetting ($F$), which measures the extent that our model forgets what it has learned in the past.  After every training step $t$ we evaluate on a test set for each of the $t$ training domains observed up to that point.  At the end of training this results an evaluation matrix $R \in \mathbb{R}^{T\times T}$, where $R_{i,j}$ measures the model's Recall@1 on test set $j$ at training step $i$.  mR@1 and $F$ are now calculated using $R$ as follows:
\begin{equation}
    \label{eq:mR@1}
    \text{mR@1} = \frac{1}{T}\sum_{t=1}^{T}R_{T,t}
\end{equation}
\begin{equation}
    \label{eq:forgetting}
    F = \frac{1}{T-1}\sum_{t=1}^{T-1}\max_{l\in 1 ... t}\left\{R_{l,t}\right\} - R_{T,t}
\end{equation}

A higher mR@1 reflects a stronger performance over all training domains, whereas a low value for $F$ indicates stronger performance on previous training domains.

\subsection{Baselines} 
\label{sec:comparison}
We evaluate the performance of our method \textit{InCloud} against the following baselines.

\textbf{Fine-Tuning (FT)}: The network is sequentially fine-tuned on all the new environments (domains) using a standard metric learning approach (\ie{}  without incremental learning). FT defines the a theoretical lower bound for incremental learning performance. 

\textbf{Joint}: The network is trained offline on all available dataset in a single training step. Joint training is regarded as the upper bound performance for an incrementally trained network.

\textbf{LwF}: Learning without Forgetting (LwF) \cite{Li2018LearningWF}, a well-established incremental benchmark which performs knowledge distillation between output logits across training steps using Kullback–Leibler divergence.

\textbf{EWC}: Elastic Weight Consolidation (EWC) \cite{Kirkpatrick2017OvercomingCF}, another well-established incremental learning benchmark which uses the Fisher Information Matrix to identify key parameters for task performance and penalise change to these parameters across training steps.

%% file: chapters/results.tex
\section{Results}
\begin{figure*}[t]
     \centering
     \begin{subfigure}[b]{0.25\textwidth}
         \centering
         \includegraphics[width=\textwidth]{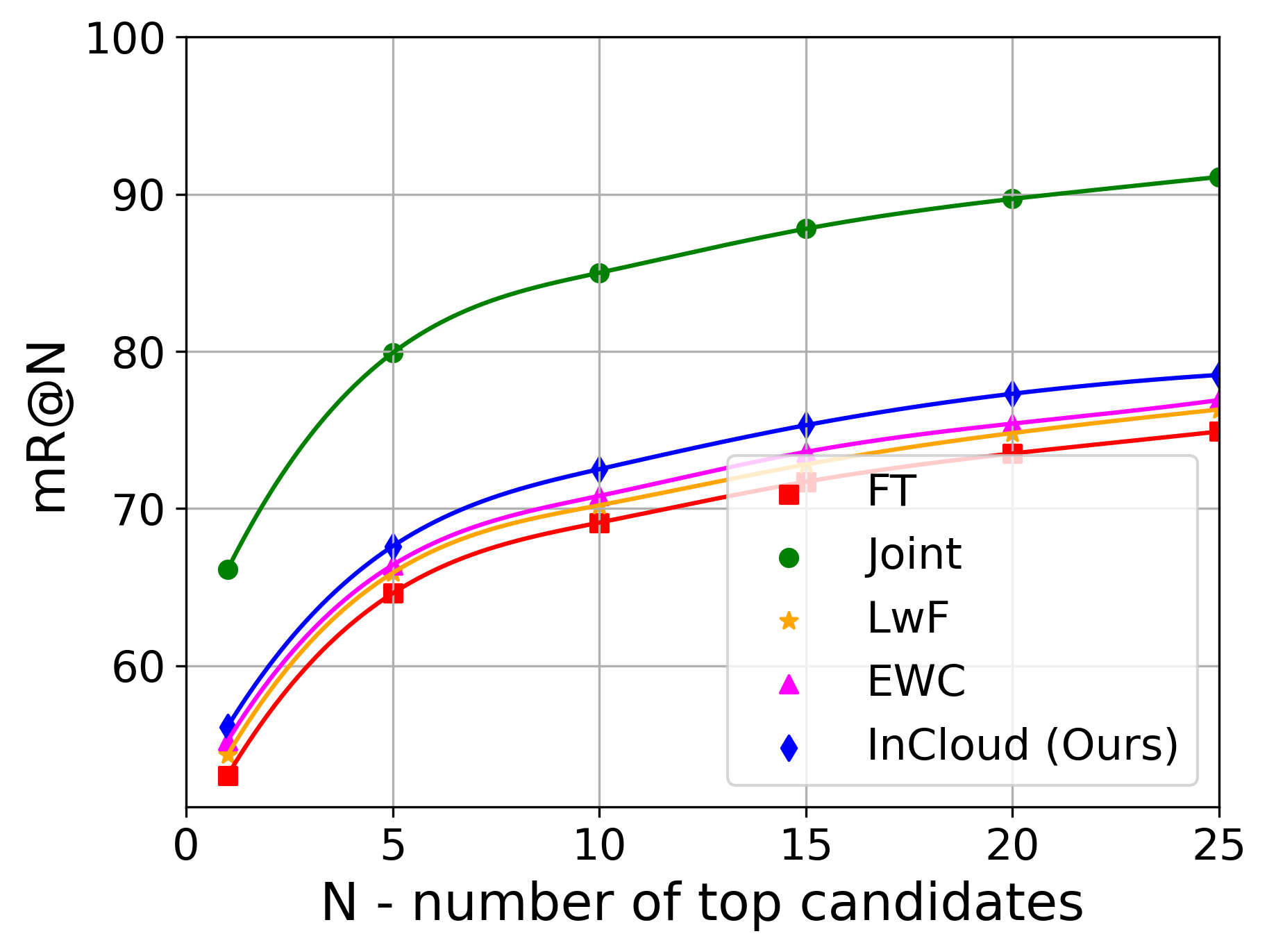}
         \caption{  PointNetVLAD \cite{Uy2018PointNetVLADDP}}
     \end{subfigure}
     \hfill
     \begin{subfigure}[b]{0.25\textwidth}
         \centering
         \includegraphics[width=\textwidth]{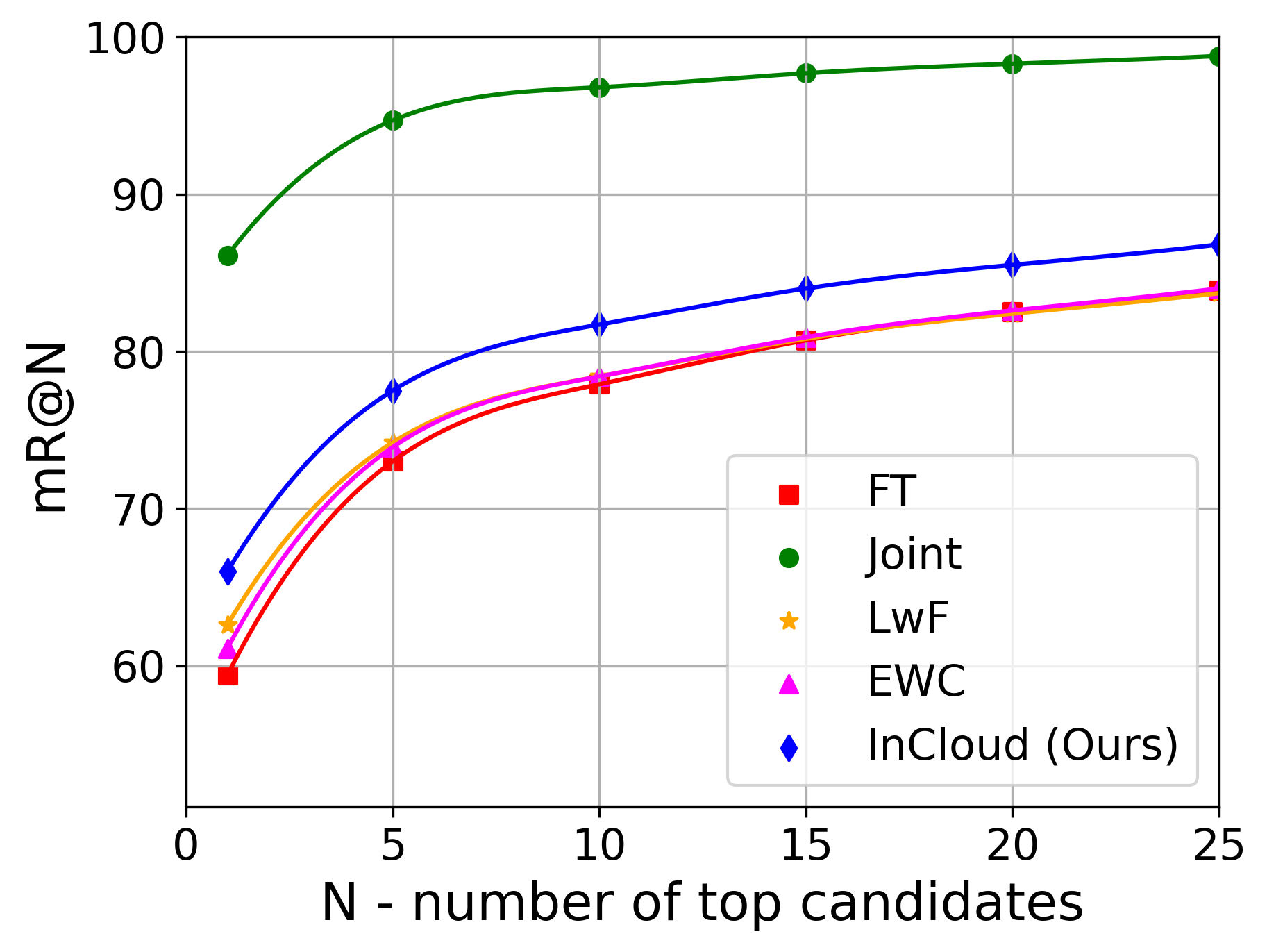}
         \caption{  LoGG3D-Net$\dagger{}$ \cite{vidanapathirana2021logg3d}}
     \end{subfigure}
     \hfill
     \begin{subfigure}[b]{0.25\textwidth}
         \centering
         \includegraphics[width=\textwidth]{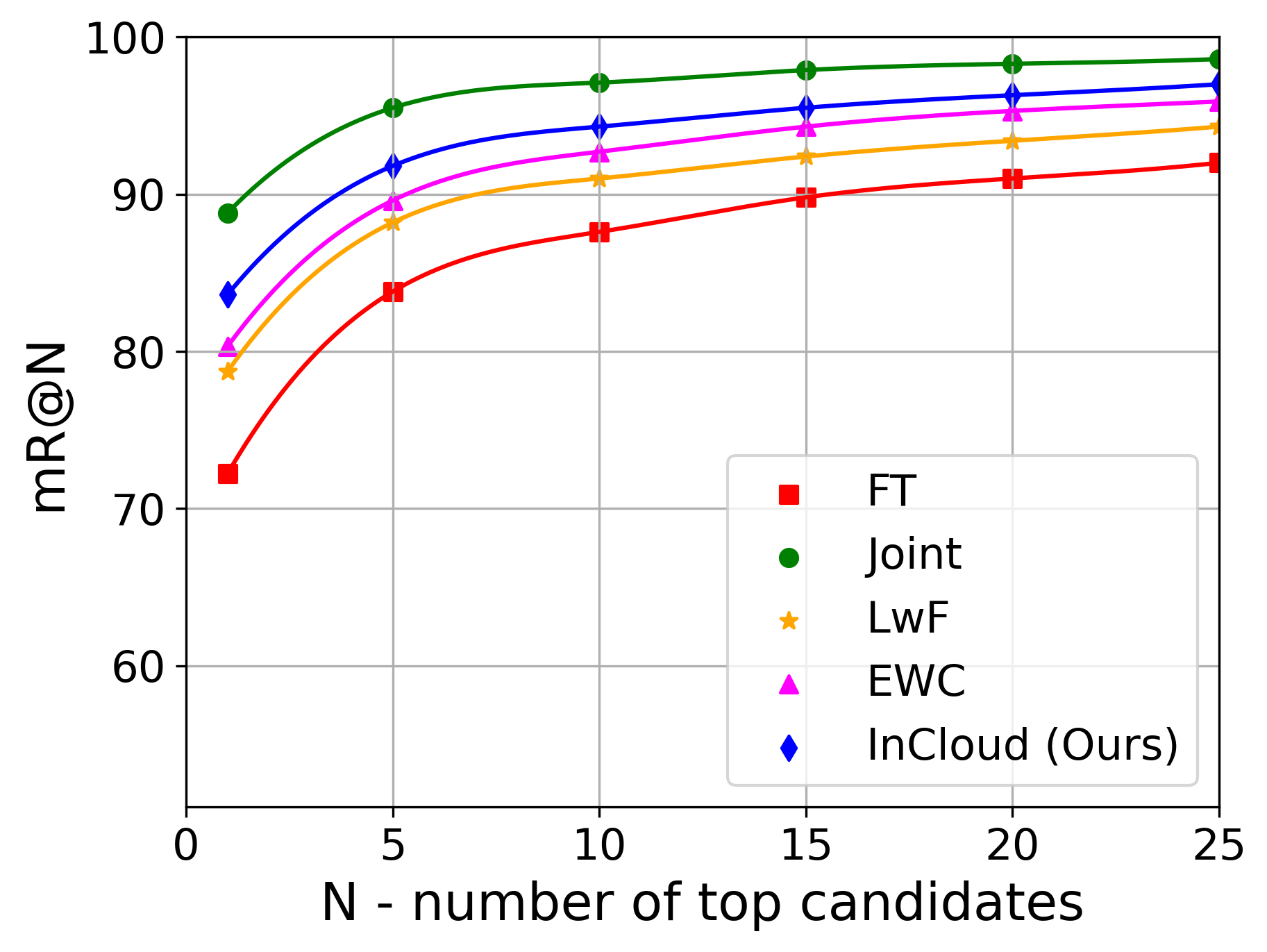}
         \caption{  MinkLoc3D \cite{komorowski2021minkloc3d}}
     \end{subfigure}

        \caption{  mR@N for different approaches trained under the \textit{4-Step} incremental protocol on (a) PointNetVLAD \cite{Uy2018PointNetVLADDP} (b) LoGG3D-Net$\dagger{}$~\cite{vidanapathirana2021logg3d} and (c) MinkLoc3D \cite{komorowski2021minkloc3d}.  As we increase the number of candidates considered in calculating the Recall, \textit{InCloud} continues to outperform competing methods.} %
        \label{fig:recall}
        \vspace{-5mm}
\end{figure*}

We evaluate \textit{InCloud} on the  \textit{2-Step},  \textit{4-Step} and  \textit{Zero-Shot} protocols outlined in Section \ref{sec:datasets}, detailing the impact of catastrophic forgetting in point cloud place recognition and that our approach is agnostic to the network architecture being used.  Additionally, we justify our approach through ablation studies on the impact of \textit{InCloud}'s components and the effect of using lower-order structural constraints in the knowledge distillation loss.

\begin{table}[t]
    \centering
    \begin{tabularx}{\linewidth}{L{0.28\linewidth}XX|XX}
        \hline 
         \vspace{-2.5mm} \multirow{3}{4em}{Method} & \multicolumn{2}{c|}{\textit{2-Step}} &  \multicolumn{2}{c}{\textit{4-Step}}   \\
         & mR@1 $\uparrow$ & F $\downarrow$ & mR@1 $\uparrow$ & F $\downarrow$ \\
         \hline 
         \textbf{PointNetVLAD}~\cite{Uy2018PointNetVLADDP} & &  &  &  \\
         FT (Lower bound) & 58.4 & 19.0 & 53.0 & 19.1 \\
          Joint (Upper bound) & 66.1 & - & 66.1 & - \\ \hdashline
          +LwF~\cite{Li2018LearningWF} & 59.1 & 16.9 & 54.3 & 17.8 \\
          +EWC\cite{Kirkpatrick2017OvercomingCF} & 58.8 & 16.1 & 55.2 & 16.8 \\
         +\textit{InCloud }(ours) & \textbf{59.7} & \textbf{13.3} & \textbf{56.1} & \textbf{8.3} \\
         \hline 
         
         \textbf{LoGG3D-Net$\dagger{}$}~\cite{vidanapathirana2021logg3d}  & &  &  &  \\
          FT (Lower bound) & 71.6 & 20.1 & 59.4 & 22.6 \\
         Joint (Upper bound) & 86.0 & - & 86.0 & -  \\ \hdashline
         +LwF~\cite{Li2018LearningWF} & 73.2 & 16.3 & 62.6 & 17.7 \\
         +EWC\cite{Kirkpatrick2017OvercomingCF} & 72.6 & 17.0 & 61.1 & 21.9 \\
         +\textit{InCloud }(ours) & \textbf{73.9} & \textbf{13.1} & \textbf{66.0} & \textbf{12.4} \\
         \hline 
         \textbf{MinkLoc3D}\cite{komorowski2021minkloc3d} & &  &  &  \\
         FT (Lower bound) & 83.6 & 20.1 & 72.2 & 22.4\\
         Joint (Upper bound) & 92.7 & - & 92.7 & -\\  \hdashline
         +LwF~\cite{Li2018LearningWF} & 84.9 & 11.3 & 78.2 & 6.1 \\
         +EWC\cite{Kirkpatrick2017OvercomingCF} & 85.8 & 7.5 & 80.3 & 9.7 \\
         +\textit{InCloud }(ours) & \textbf{87.7} & \textbf{5.2} & \textbf{83.6} & \textbf{5.7} \\
         \hline 
    \end{tabularx}
        \caption{  Incremental learning results for networks trained on the \textit{\textit{2-Step}} and \textit{4-Step} incremental protocols.}
        \label{tab:seen_env}
    \vspace{-6mm}
\end{table}

\subsection{Catastrophic Forgetting}
In this section we investigate the impact of catastrophic forgetting for point cloud place recognition.  \figref{fig:forgetting} illustrates the impact of catastrophic forgetting in the \textit{4-Step} protocol.  At each training step the network overfits to a new training environment, optimising performance on that domain.  However as new environments are introduced performance on old domains degrades due to catastrophic forgetting, with drops of up to $34\%$, $31.5\%$ and $9.7\%$ experienced on the Oxford, DCC and Riverside environments, respectively.  Forgetting is smaller on environments visited later in training, as less training steps have been taken by the network since their introduction.  These results indicate the severity of impact catastrophic forgetting can have when learning from new domains in point cloud place recognition, and the importance of introducing methods to alleviate this issue.

\subsection{Performance on 2-Step and 4-Step Protocols}

Table \ref{tab:seen_env} reports results for the \textit{2-Step} and \textit{4-Step} incremental protocols.  As expected the \textit{4-Step} protocol experiences a higher drop in mR@1 and greater forgetting than the \textit{2-Step} protocol.  We attribute this to the network more directly overfitting to the training distribution when the new training sets are significantly narrower in domain, which in turn impacts overall generalisability and increases forgetting.  On the \textit{4-Step} protocol \textit{InCloud} demonstrates significant improvement over the competing baselines, with a $3.1\%$, $6.6\%$ and $11.4\%$ improvement in mR@1 and a $10.8\%$, $10.2\%$ and $16.7\%$ reduction in forgetting on PointNetVLAD \cite{Uy2018PointNetVLADDP}, LoGG3D-Net$\dagger{}$~\cite{vidanapathirana2021logg3d} and MinkLoc3D~\cite{komorowski2021minkloc3d}, respectively, over the FT and outperforming both LwF \cite{Li2018LearningWF} and EWC \cite{Kirkpatrick2017OvercomingCF} across all metrics and architectures.  

On the \textit{2-Step} protocol \textit{InCloud}'s gains in mR@1 are more marginal with a $1.3\%$, $1.0\%$ and $4.1\%$ improvement across the three architectures.  However \textit{InCloud} still significantly outperforms the competing baselines in preserving performance on previously visited domains: we demonstrate $5.7\%$, $3.2\%$ and $14.9\%$ reduction in forgetting across the three architectures compared to the FT , and once again outperform both LwF \cite{Li2018LearningWF} and EWC \cite{Kirkpatrick2017OvercomingCF}.

Finally, \figref{fig:recall} presents mean Recall@N for a range of  top candidates numbers N for the three network architectures.  As we increase the number of top candidates \textit{InCloud} continues to outperform competing methods, though there is still a significant gap between \textit{InCloud} and the Joint training upper bound for PointNetVLAD \cite{Uy2018PointNetVLADDP} and LoGG3D-Net$\dagger{}$\cite{vidanapathirana2021logg3d}.  These results demonstrate the significant benefits of \textit{InCloud} across a variety of incremental protocols and network architectures.

\subsection{Performance on Zero-Shot Protocol}

\begin{table}[t]
    \centering
    \begin{tabularx}{\linewidth}{m{0.25\linewidth}M{0.2\linewidth}M{0.15\linewidth}X}
        \hline 
         Method & \textbf{PointNet VLAD}~\cite{Uy2018PointNetVLADDP} & \textbf{LoGG3D-Net$\dagger{}$}~\cite{vidanapathirana2021logg3d}  & \textbf{MinkLoc3D}~\cite{komorowski2021minkloc3d}\\\hline
         Non-Incremental & \multicolumn{1}{c}{77.9} & \multicolumn{1}{c}{73.7} & \multicolumn{1}{c}{89.6} \\ \hdashline
         +LwF \cite{Li2018LearningWF} & \multicolumn{1}{c}{83.4} & \multicolumn{1}{c}{80.7} & \multicolumn{1}{c}{88.2} \\
         +EWC \cite{Kirkpatrick2017OvercomingCF} & \multicolumn{1}{c}{\textbf{86.9}} & \multicolumn{1}{c}{73.9} & \multicolumn{1}{c}{88.9} \\
         +\textit{InCloud} (ours) & \multicolumn{1}{c}{85.3} & \multicolumn{1}{c}{\textbf{84.8}} & \multicolumn{1}{c}{\textbf{90.7}} \\\hline

    \end{tabularx}
    \caption{  \textit{Zero-Shot} results on KITTI-00.  Reported values are mean Recall@1.}
    \label{tab:unseen_env}
    \vspace{-5mm}
\end{table}

\tabref{tab:unseen_env} reports results on the KITTI dataset under the \textit{Zero-Shot} protocol outlined in \secref{sec:datasets}.  Since the mR@1 and Forgetting metrics %
are not applicable in a zero-shot setting, we report Recall@1 instead.  We compare the non-incremental learning result - that is, a network is fine-tuned sequentially over 4 training steps without any incremental learning compensation and 
- with LwF \cite{Li2018LearningWF}, EWC \cite{Kirkpatrick2017OvercomingCF} and \textit{InCloud} on all three network architectures.

\textit{InCloud} improves over the non-incremental approach by $7.4\%$, $11.1\%$ and $1.1\%$ on KITTI-00, and outperforms LwF \cite{Li2018LearningWF} across all architectures and EWC \cite{Kirkpatrick2017OvercomingCF} on LoGG3D-Net$\dagger{}$ \cite{vidanapathirana2021logg3d} and MinkLoc3D \cite{komorowski2021minkloc3d}.  These results support our hypothesis that by preventing overfitting at each training step and forcing the network to maintain a more general embedding, \textit{InCloud} significantly improves the forward-transferability of our global representation to previously unseen domains.

\subsection{Effect of Distillation Losses}

In Section \ref{sec:lsa} we stress the importance of capturing the structural content of the embedding space when distilling knowledge between training domains.  Table \ref{tab:higher-order} compares our approach - distilling the angular relationship between vectors of a triplet - with two conventional approaches, distilling the absolute position of an embedding and the Euclidean distance between pairs of embeddings.  Compared to the absolute and Euclidean distance distillation losses, our approach reports a $2.8\%$ and $1.1\%$ improvement in mR@1 and a $1\%$ and $0.4\%$ reduction in forgetting, respectively.  This demonstrates the advantages of capturing the higher-order structure of the embedding in distillation loss while retaining the flexibility to adapt to new domains.

\subsection{Ablation Study}
\begin{table}[t]
    \centering
    \begin{tabular}{lcc}
        \hline 
         Method & mR@1 $\uparrow$ & F $\downarrow$  \\
         \hline 
         Point & 80.8 & 7.1 \\
         Euclidean distance & 82.5 & 6.1\\
         Angular  & \textbf{83.6} & \textbf{5.7} \\
         \hline 
    \end{tabular}
    \caption{  Comparison of different distillation losses.  Results are reported on the \textit{4-Step} incremental protocol using the MinkLoc3D architecture.}
    \label{tab:higher-order}
\end{table}
    \begin{table}[t]
        \centering
        \begin{tabular}{ccccc}
            \hline 
            Memory & SA & Relaxation & mR@1 $\uparrow$ & F $\downarrow$ \\
            \hline 
            - & - & - & 72.2 & 22.4  \\ 
            \cmark & - & - & 77.3 & 15.4  \\ 
            \cmark & \cmark & - & 80.1 & \textbf{3.0}  \\
             - & \cmark & {\cmark} & {81.0} & {9.9} \\
            \cmark & \cmark & \cmark & \textbf{83.6} & 5.7 \\
            
            \hline 
        \end{tabular}
        \caption{  The effect of different components of the loss in the proposed method. Results are reported on the \textit{4-Step} incremental protocol using the MinkLoc3D architecture.}
        \label{tab:components}
        \vspace{-5mm}
    \end{table}
    
Table \ref{tab:components} provides a breakdown of how the different components of our approach impacted the overall performance of the network.  Each major component of \textit{InCloud} significantly improves the mR@1; however, adding the relaxation also results in a small increase in forgetting.  This result is indicative of the nature of the plasticity-stability trade-off in incremental learning protocols; though the distillation relaxation is overall advantageous in allowing the network to adapt more smoothly to new domains, making it more \textit{plastic}, by relaxing the distillation loss we sacrifice a small amount of \textit{stability} as a result.  %

%% file: chapters/conclusion.tex
\section{Conclusion}
In this paper, we explore the novel challenge of incremental learning in the context of point cloud place recognition and introduce a new approach, \textit{InCloud}.  \textit{InCloud} distills the angular relationship between global representations to preserve the complex structure of the embedding space between incremental training steps, and employs a new distillation relaxation approach which gradually tempers the distillation as training progresses.  To evaluate our approach, we introduce new benchmarks on four LiDAR datasets of Oxford, MulRan, In-house, and KITTI and train our approach on multiple state-of-the-art network architectures.  We show that agnostic to the network architecture \textit{InCloud} effectively addresses the problem of catastrophic forgetting for point cloud place recognition, allowing for models to be updated on new domains without the restriction of retaining and re-training the network jointly on all legacy data.